# A Critical Study on Tea Leaf Disease Detection using Deep Learning Techniques

Nabajyoti Borah[1], Raju Moni Borah[2], Bandan Boruah[3], Purnendu Bikash Acharjee[4], Sajal Saha[5] and Ripjyoti Hazarika[6]

[1,2,3,6]Student of Kaziranga University
[4]Associate Professor School of Computing Sciences, Kaziranga University
[5]Director product and innovation and professor, Dept of Computer Science and Engineering, Adamas university

[1]sirnaba@gmail.com,
[2]raju42fun@gmail.com,
[3]bandanboruah1234@gmail.in,
[4]pbacharyaa@gmail.com,
[5]sajal1.saha@adamasuniversity.ac.in,
[6]ripjyotihazarika123@gmail.com

**Abstract**

The proposed solution is Deep Learning Technique that will be able classify three types of tea leaves diseases from which two diseases are caused by the pests and one due to pathogens (infectious organisms) and environmental conditions and also show the area damaged by a disease in leaves. Namely Red Rust, Helopeltis and Red spider mite respectively. In this paper we have evaluated two models namely SSD MobileNet V2 and Faster R-CNN ResNet50 V1 for the object detection. The SSD MobileNet V2 gave precision of 0.209 for IOU range of 0.50:0.95 with recall of 0.02 on IOU 0.50:0.95 and final mAP of 20.9%. While Faster R-CNN ResNet50 V1 has precision of 0.252 on IOU range of 0.50:0.95 and recall of 0.044 on IOU of 0.50:0.95 with a mAP of 25%, which is better than SSD. Also used Mask R-CNN for Object Instance Segmentation where we have implemented our custom method to calculate the damaged diseased portion of leaves.

***Keywords:*** Tea Leaf Disease, Deep Learning, Red Rust, Helopeltis and Red Spider Mite, SSD MobileNet V2, Faster R-CNN ResNet50 V1 and Mask RCNN.

## 1. Introduction

Tea plays an important role in the field of Human life. One of the popular beverages across the world is tea, and the world's second largest producer of tea is India. Proper nurturing is required for healthy growth of tea plants. Tea plant diseases are a threat in the production of tea and identifying the disease remains difficult. Insects, mites, fungi, bacteria and algae are the most common causes of tea plant diseases.

So, we came with an idea to provide an image-based automatic inspection interface. The proposed solution uses a Deep learning CNN algorithm. Our proposed solution would classify three types of diseased leaves, Red spider mite Helopeltis and Red rust. It's very important to remove the diseased leaves because it can be further spread to healthy leaves causing reduction in tea production. Red Spider Mite-affected leaves develop reddish brown markings on the upper surface, which turn red in extreme infestations, inhibiting the plant's photosynthesis. If the temperature rises, so does the level of infestation. The Tea Mosquito Bug(Helopeltis) only attacks the young shoots that make up the actual crop of tea. Changes in color are observed in affected leaves, translucent and light brown. Leaves may even curl-up. Helopeltis can be found on tea bushes almost all year, but in Northeast India, the occurrence peaks in June and July, sometimes stretching into September, when the number of rainy days is at its highest. Algae causes Red Rust, which appears as orange-yellow irregular spots on the leaf's upper surface. Wind or water splashing back up into the leaves spreads the disease. 4-8 hours of low light intensity, warm temperatures, and moisture humidity, such as dew or fog, are favorable to rust diseases, accompanied by 8-16 hours of high light intensity, high temperatures, and slow drying of leaf surfaces.

The following is how the rest of the paper is structured: A quick summary of the currently known conventional DL methods for plant disease detection can be found in Section 2. Section 3 provides methodology of the proposed system. The conclusions are presented in Section 4, Section 5 provides Acknowledgement followed by a list of references.

## 3. Related Works

As the development in the field of artificial intelligence, deep learning. It has been widely utilized in the field of agriculture to identify plant diseases. The System [1] proposed in presents a tea leaf disease recognizer that recognizes affected leaves of the tea plants. The input of the proposed system is the images of leaves of tea taken by the digital camera. The proposed system combines feature extraction and neural network ensemble for training. The captured images are processed and after feature extraction the extracted features are used as input which identifies tea leaf diseases. The proposed solution has 91 % of accuracy.The system proposed [2] used Support

Vector Machine (SVM) to recognize the diseases in the leaves. Features were analyzed during the classification. When a new picture is uploaded into the system the most suitable match is found and the disease is recognized. The approach retains an accuracy of more than 90%. In the proposed system [3] they selected three models for their tea leaf disease classification namely CNNs model algorithm identified tea leaf diseases most accurately, with an average classification accuracy of 90.16%, the SVM algorithm was 60.62% and the MLP algorithm was 70.77%. In the proposed system [4] They used CNN model for classifying the image of tea plant diseases with accuracy of 89.64% and for pre-trained they used GoogLeNet, Xception, and Inception-ResNet-v2 architecture. The accuracy of GoogLeNet architecture was 76.65%, Xception architecture was 87.10% and Inception-ResNet-v2 was 87.95%. In the proposed system [5] here CNN model is used to recognize the image of tea leaf diseases an in CNN is 93.75%, while the accuracy of SVM and BP neural networks is 89.36% and 87.69% respectively. A total of 25186 images were obtained through data enhancement. Then 25186 images are used as a training data set, and the rest 2858 images are used as test sets. The proposed system [6] is an enhanced RCNN model that is integrated with ResNet CNN based on which an attempt was made to detect and classify the crop diseases at the early stages and enhance the productivity of the crop. The proposed model used dataset of 44055 images of different plants along with the 32 classes of the diseases are considered. In the proposed system [7] They utilised the CNN model to categorise the images. The implemented models were trained on an open dataset that included 14 plant species, 38 categorical illness classifications, and healthy plant leaves. Using InceptionV3, InceptionResNetV2, and MobileNetV2, the implemented models obtained illness classification accuracy rates of 98.42 percent, 99.11 percent, 97.02 percent, and 99.56 percent, respectively. In the proposed system [8] they took advantage of Faster R-CNN and Mask R-CNN are two distinct models utilised in these approaches, with Faster R-CNN used to identify the kinds of tomato illnesses and Mask R-CNN used to detect and segment the locations and shapes of diseased regions. In the Faster R-CNN, the mAP value of ResNet-101 is the highest, reaching 88.53%. VGG-16 with mAP of 86.09%, ResNet-50 with mAP of 88.41%.and MobileNet with mAP of 88.39%. In the Mask R-CNN model, ResNet-101 also performed well, with a mAP value as high as 99.64%. In [9] the authors created a new dataset with 79,265 images, which contained images of leaves in real surroundings, at different angles, and in various weather conditions, labelled both for classification and detection tasks. The dataset is more comprehensive, which improves the classification accuracy and practical applicability of the model. Augmentation techniques were used. They trained the model and achieved an accuracy of 93.67%. In [10] the authors used three types of deep learning techniques for their evaluation purpose SSD, Faster RCNN, RCFN. They selected, the dataset which consists of images of 14 plant varieties. The SSD model trained with an Adam optimizer exhibited the highest mean average precision (mAP) of 73.07%. The successful identification of 26 different types of defected and 12 types of healthy leaves. In the [11] The recommended system has the following characteristics: MobileNet V2, a deep learning architecture, has been used to identify three kinds of tomato illnesses. To fulfil the input criteria of the MobileNet V2 model, the input picture size is rescaled to 224*224 pixels. The method has been assessed. With better than 90% accuracy, MobileNet V2 can diagnose the disease.

**Table 1 Related work**

| Paper | Data size | Preprocessing Data size | Best CNN architecture | Transfer Learning | Accuracy |
|---|---|---|---|---|---|
| [3] | 3810 | 7905 | Modified AlexNet | yes | 90.16% |
| [6] | 44055 | N/A | Faster-RCNN with ResNet | yes | 90.6% |
| [8] | 286 | 1,430 | ResNet-101 | yes | 88.53%. |
| [10] | 54,309 | N/A | SSD | yes | 73.07% |
| [11] | 4,671 | N/A | MobileNet V2 | yes | 90% |

## 4. METHODOLOGY

A. Dataset collection and Pre Training Process Augmentation
Tea leaves have variant diseases depending on the temperature, soil, pests etc. we collected mainly Red Rust in December, Helopeltis and Red Spider Mite in the month of January to March. We collected the images from tea gardens in Jorhat namely Assam Agriculture University Tea Garden Mohbondha Tea garden and Socklating Tea Estate. The images were taken using mobile devices with high pixel quality of 4000x3000px width and height without any flash. The images are taken in daylight. Images are taken at close range to capture detailed images of diseases on the field. For each class we collected 500 images, a total 1500 of all three diseases. All the images are verified by Dr. Popy Bora(Scientist) and Dr B.C Nath(Assistant professor) from Assam Agriculture University.

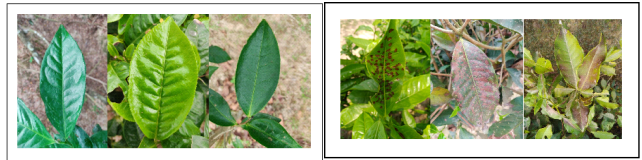
Figure 1 Helopeltis, Red Rust and Red Spider Mite

Since the images are too low we used augmentation to increase the images. Augmentation is a technique used to increase the total dataset size if data is low. It uses techniques like rotation, zoom, crop to get many different variations for a single image. We for our use case used rotation and crop. Rotation to get multi angle images of tea leaves, so that images taken on any angle can be detected. Random crop is used to enlarge the image disease parts which are very small in some cases and harder for training. So after augmentation we have a total 4500 images, 1500 original images, 1500 randomly rotated images, 1500 randomly crop images.

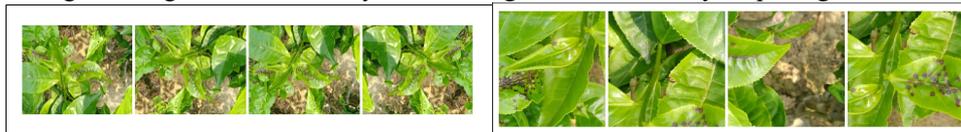
Figure 2: Augmentation Random Rotation and Augmentation Random Crop

B. OBJECT DETECTION MODEL

a)Image annotation and TF Record: After all the images are verified, labeling is being done manually. Each disease is classified in the image and a bounding box is created around it with a label as the disease name. If a single image contains multiple diseases in it then all are labeled properly. For red rust the boxes are created grouping more than one Red Rust Spots and same for Helopeltis. In some images single Helopeltis are also labeled. While labeling for red spider bite the area covered with the red part is labeled using different box sizes.

While labeling the diseases in the leaves we must include some of the outer part of the disease, like the green part of the leaf so that the model can get the edges of the disease part properly.

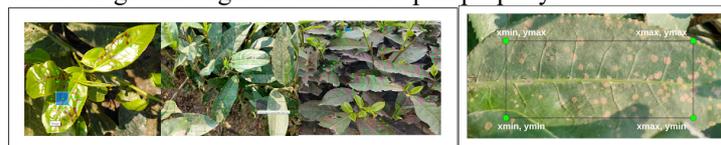
Figure 3: Disease Labeling and Bounding Boxes

The tool we used to label the images LabelImg which is a free, open source tool for graphically labeling images. It's written in Python and uses QT for its graphical interface. Now after labeling an image the annotation data is stored in PASCAL/VOC format, it is a XML file which contains the class label and the bounding boxes dimensions xmin, xmax, ymin, ymax.

b)TF Record format dataset : Since tensorflow recommends storing and reading the data in tfrecord files. Because with the help of tfrecord files we can shard(divide) the dataset into multiple shards so that it can be trained on multiple TPU or GPU. Since the data stored in binary format in tfrecord files it impacts performance at the time training the model also, because tfrecord files take less space, faster to read and write from a HDD, which is slower than a SSD and load the data in memory in batches instead of whole dataset, which helps less memory devices. Tfrecord file is a sequence of byte-string, so first with the help of the XML file each image is converted into a numpy array and then converted into a byte-string with all the bounding boxes and class labels in the XML file. From the total dataset 150 images are used for evaluation and 1350 are used for training the model. A test.tfrecord of 150 images is created and train.tfrecord for 1350 images is created.

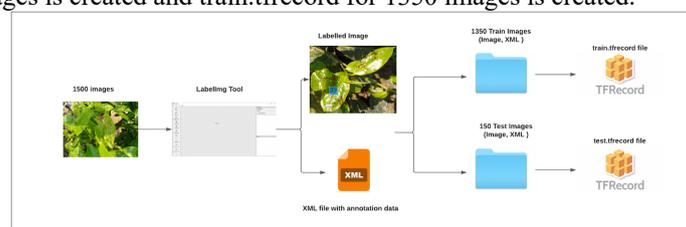
Figure 4: Workflow of Labelling and TFrecord

c)Tensorflow model zoo 2: The models used for the task are from TensorFlow 2 Detection Model Zoo. The model zoo is a collection of object detection models which are pre-trained on the COCO 2017 dataset[12]. COCO is a large-scale object detection, segmentation dataset. It has 330,000 images of 80 object categories. Models from the model zoo can be downloaded as a zip file, which contains the saved freeze model, which is a pre-trained saved model on coco dataset and can be directly used for prediction on the trained categories. Second important thing is the saved checkpoint, this file contains the pre-trained features of the coco-dataset images. Instead of training the whole object detection model from scratch we will use these features as a part of transfer learning. And lastly we have a pipeline config file of the whole model which consists of the hyperparameters, augmentation, batch size, number of classes, feature layer, loss etc. We can tune these parameters according to our needs to train our model for our own tea leaves disease dataset. While selecting models from the model zoo there are two main metrics we have to consider, they are also given in the model zoo with the model. Thats speed and accuracy (mAP). Speed metrics is used to measure the speed at which a model detects an object in the image, unit of the speed is (ms). The second metric is the accuracy or COCO mAP, it is called COCO mAP because it is calculated on the coco dataset. Mean Average Precision (mAP) or Average Precision is used to evaluate the model after training on a custom dataset. To calculate the mAP we first need precision and recall score of the model. Precision measures the "false positive rate" or we can say the ratio of true object detection to the number of total objects in the dataset. Recall is calculated as the ratio between the total number of positive objects correctly classified as positive to the total number of positive samples.

Precision = True positive / (true positive + false positive)

Recall = True positive / (true positive + false negative)

Now in object detection models to calculate precision and recall we use IOU (intersection over union) values. IOU is calculated using the ground truth box (self labelled box) and the prediction box (model prediction box).

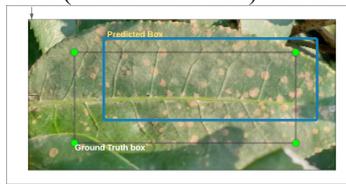

**Figure 5: Ground truth and prediction box**

To calculate the IOU we need the intersection area and the union area between the predicted box and ground truth box.

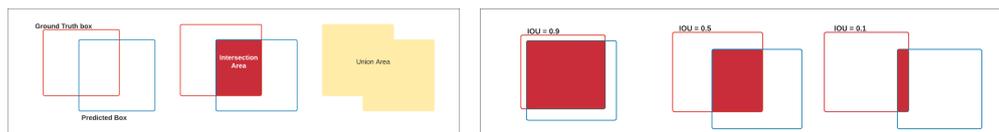

**Figure 6: Intersection area and union area and Figure 10: IOU score**

IOU = Area of Intersection / Area of Union

We use different threshold values to judge a model whether it is able to predict the box location or not. If the IOU score is within or greater than the threshold value, it means that the intersection area is high between the predicted and ground truth box. Suppose the threshold value is 0.5 and the IOU score calculated for an object is greater than the threshold value or equal, we can say that the model predicted the correct position of the object in an image. With this IOU score we can now calculate the true positive and negative predictions which further helps in calculating the precision and recall. For true positive IOU > 0.5 if threshold is 0.5 else it is false positive. Now we need to calculate the mAP of the model, mAP is calculated using the AP (Average precision) of each class, so the mean of the APs for each class is the final mAP of the model. Therefore, it is called mean Average Precision in context of object detection. Average Precision for each class can be calculated using the precision and recall curve which is calculated using different threshold values. In object detection AP is the average over multiple IOU values. It is set by the COCO dataset only to evaluate a model. There are a total 10 IOU levels used from 0.5 to 0.95 with step of 0.5.

So if we go step wise step to calculate the mAP:

- Calculate the AP of each class on different IOU thresholds, and then calculate average AP for each class.
- Then calculate the mean AP by averaging the AP of different classes, and that will be the final mAP of the model.

mAP Formula:

$$mAP = \frac{1}{|classes|} \sum_{c \in classes} \frac{\#TP(c)}{\#TP(c) + \#FP(c)}$$

**Figure 7: mAP Formula**

d)Object Detection models for the project: The proposed models used are

**Table 2 Object Detection Models**

| Model | Speed(ms) | COCO mAP |
|---|---|---|
| SSD MobileNet V2 FPNLite 640x640 | 39 | 28.2 |
| Faster R-CNN ResNet50 V1 640x640 | 53 | 29.2 |

e)SSD MobileNet V2 FPNLite 640x640: is a Single-Shot multibox detection network built for object detection tasks with more speed but less accuracy. The model mobilenet architecture is suitable for low end devices due to less computation power need. With an SSD detector it can perform real time detection much faster compared to the other models in Tensorflow model zoo 2. The main change added by tensorflow here is the backbone layer of the SSD network, instead of VGG-16, they used Mobilenet V2

SSD (Single Shot Detector): SSD means it can in a single forward pass can do object localization and classification. The SSD network consists of a backbone layer which is a feature extraction layer. Now what the 6 main layers of SSD does is that it predicts 8732 boxes for a single image using default anchor boxes, then the work of the non-max suppression layer is to remove boxes using a threshold value. The threshold value is compared with IOU value and if any box IOU value less than the threshold is removed and finally 200 boxes remain for a single image.

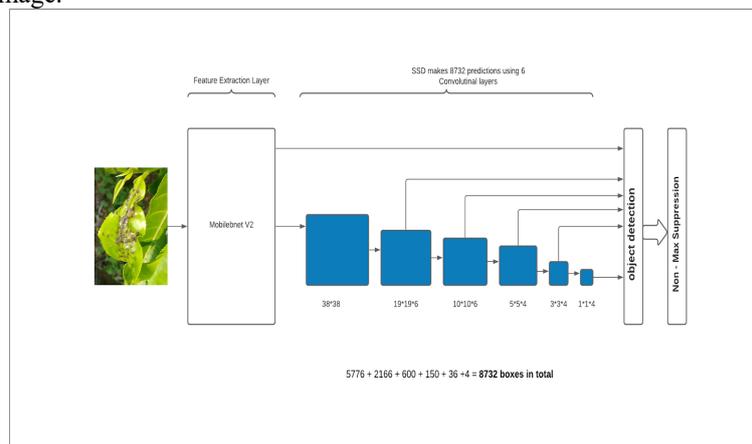

**Figure 8: SSD Network with Mobilenet V2 Backbone**

MobileNet V2: The backbone layer of the SSD network is the mobilenet V2 convolutional neural network, which is an efficient network built for next generation mobile devices. The network layers are well optimized for mobile devices to run deep learning models in mobile devices directly without any high computation needs. It uses depth-wise separable convolutions layers to build lightweight deep neural networks.

f)Faster R-CNN ResNet50 V1 640x640: is a two-stage object detection model which is divided into two sections. The first stage is aimed at creating ideas, or evaluating which areas of the image are most likely to contain objects. This is performed by using a region proposal network - RPN algorithm, that takes a feature map generated by a backbone network (ResNet50) and determines areas where an object exists.

Faster R-CNN: It [13] begins by using CNN to extract feature maps from the source images, then passes those maps through an RPN to generate object proposals. Finally, the bounding boxes are projected and these maps are classified. RPN (Region Proposel Network) is small neural network sliding on the last feature map of the convolution layers and predict whether there is an object or not and also predict the bounding box of those objects.

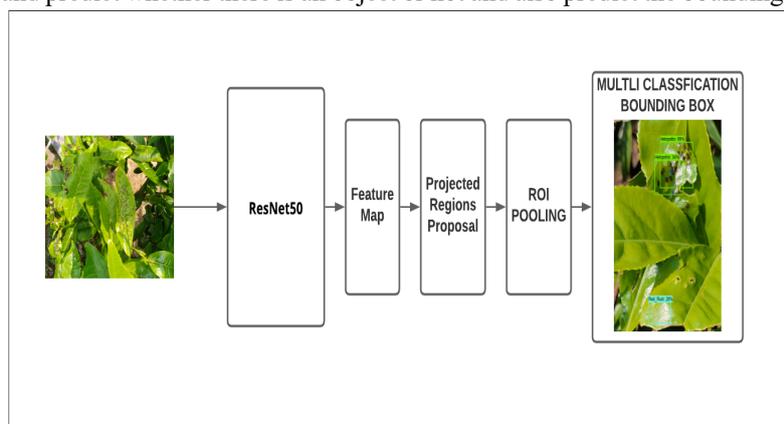

**Figure 9: Faster R-CNN with ResNet50 V1**

ResNet50 V1: consists of a convolutional neural network of 50 layers. It's a network that's been pre-trained on over a million images from ImageNet. ResNet outperforms other classification approaches by increasing the network's depth to produce a function with more semantic details. ResNet50 is used to derive features from the area proposal as well as the classification stage. It makes use of a skip link. A skip connection provides a shortcut from a shallow layer to a deep layer by connecting the input of a convolutional block to its output. This helps in resolving the vanishing gradient problem. The path and magnitude of an error gradient are determined during the training of a neural network and are used to adjust the network weights in the correct direction and by the appropriate number. In case of vanishing gradients, the loss function approach zero, making it difficult to train the network.

ImageNet: The ImageNet [14] project is a massive graphic library created to help with the development of visual object recognition applications. It consists of 14 million images & 22 thousand visual categories

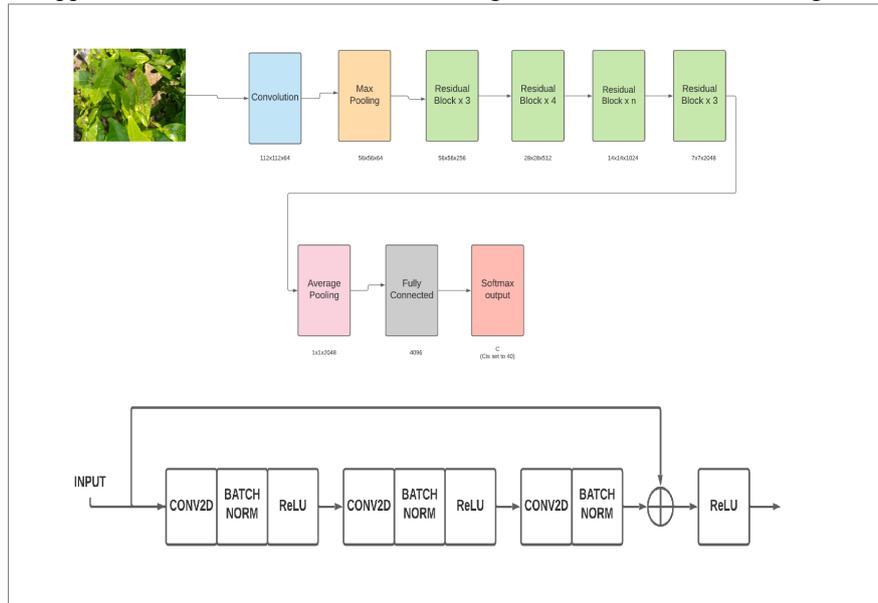

**Figure 10: ResNet50 Architecture and Residual Block**

g) Configure models for Tea Leaves Dataset: Now before starting the training of the models we first have to configure the pipeline config fickle which comes with the model zip file. For our training the params we configured are

**Table 3 Configure model**

| Parameter | SSD MobileNet V2 FPNLite | Faster R-CNN ResNet50 V1 |
|---|---|---|
| No. of classes | 3 | 3 |
| Learning rate | 0.08 | 0.04 |
| Image shape | 640, 640, 3 | 640, 640, 3 |
| Feature_extractor | ssd_mobilenet_v2_fpn_keras | faster_rcnn_resnet50_keras |
| Activation | RELU_6 | RELU_6 |
| IOU_threshold | 0.6 | 0.6 |
| Batch size | 10 | 4 |
| Augmentation | random_flip, random_crop | random_flip |
| Optimizer | Momentum optimizer | Momentum optimizer |

h) Training phase: All the models are trained for a total 5000 steps and for each 1000 steps a checkpoint is being saved to evaluate the model.

**Table 4 Data distribution for train and test**

| Leave Type | Total Images | Train Images | Test Images | Augmented Images for Training |
|---|---|---|---|---|
| Red Rust | 500 | 450 | 50 | 450 * 2 = 900 |
| Helopeltis | 500 | 450 | 50 | 900 |
| Red Spider Mite | 500 | 450 | 50 | 900 |

| Total | 1500 | 1350 | 150 | 2700 |

**Table 5 Training Log Data**

| Model | Speed per step | Ram Used% | GPU Used % | Steps Trained for |
|---|---|---|---|---|
| SSD MobileNet V2 FPNLite 640x640 | 2.90 s | 63.58 | 31.25 | 10000 |
| Faster R-CNN ResNet50 V1 640x640 | 0.921 s | 66.66 | 61.81 | 10000 |

i) Model Evaluation

**Table 6 Model Evaluation Results**

| Model | Average Precision(IOU= 0.50:0.95) | Average Recall(IOU= 0.50:0.95) | mAP |
|---|---|---|---|
| *Faster R-CNN ResNet50 V1 640x640* | 0.252 | 0.044 | 0.2521 |
| *SSD MobileNet V2 FPNLite 640x640* | 0.209 | 0.02 | 0.20 |

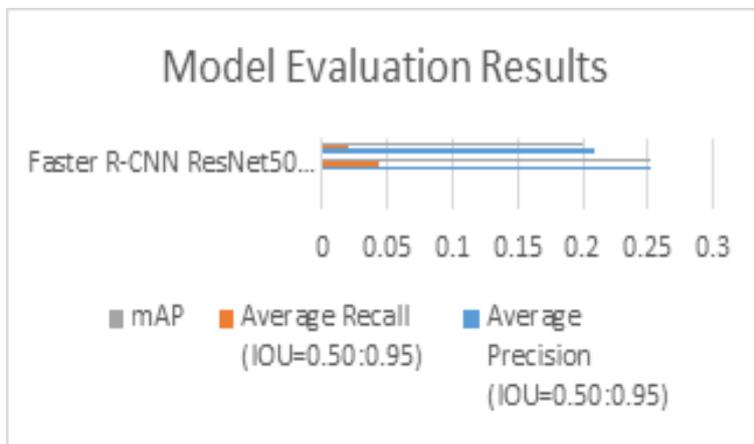

**Figure 11: Model Evaluation Results**

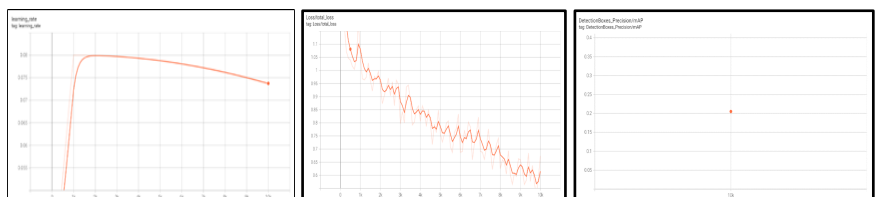

**Figure 12: SSD learning rate, SSD total loss and SSD precision**

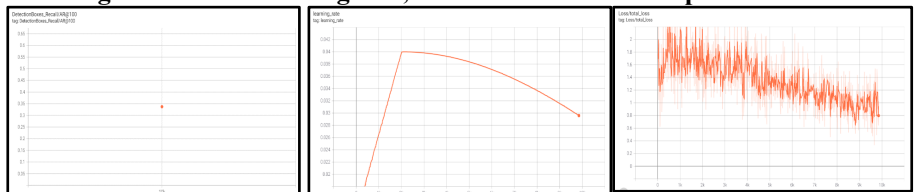

**Figure 13: SSD recall, Faster RCNN learning rate and Faster RCNN total loss**

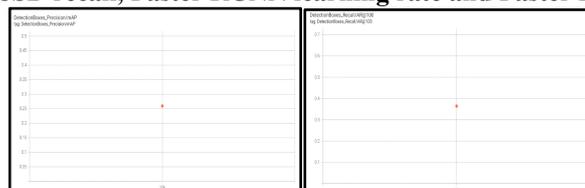

**Figure 14: Faster RCNN precision and Faster RCNN recall**

Here the main evaluation metrics is mAP (mean average precision), The Mean Average Precision (mAP) is the average AP over multiple IoU thresholds for each class. For example, mAP@[0.5:0.05:0.95] corresponds to the AP for IoU ratio values ranging from 0.5 to 0.95, at intervals of 0.05, averaged over all classes. The final result of both the models trained for 10000 steps show that Faster R-CNN has 25% mAP and the SSD has 20%.

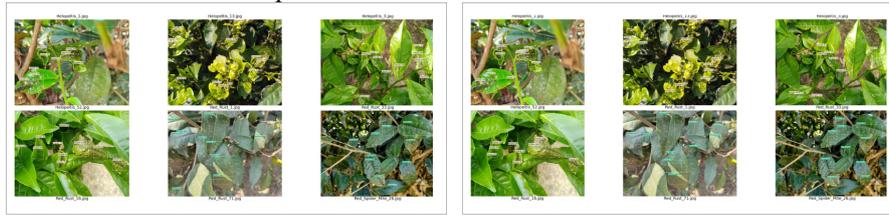

**Figure 15: SSD Mobilenet V2 predictions and Faster R-CNN predictions**

We can see that the faster rcnn model is able to classify the images with an average score of 80% to 90%, while SSD remains at 30% - 50% score.

C. INSTANCE SEGMENTATION

a)Model: The model we used for instance segmentation is Mask R-CNN Inception ResNet V2, the model developed above Faster R-CNN. It used Resnet101 and Resnet50 for feature extraction, R-CNN part for object detection and last masking of the object. The model we trained used Resnet101.

b)Data preparation: Now for annotation of tea leaf images we used labelme, it can label images in polygon shape also, which is needed for Mask R-CNN model. Each image contains diseased leaf and also healthy leaf, but we labelled the diseases part of the leaf and the leaf itself, so that model can segment the leaf and the disease inside it also. The disease which are stick to each other are labelled in single polygon. The purpose of annotating both leaves and the disease is because we want to find the mask of both leaf and the disease, so that we can further find out the area of damage inside that leaf. Now the labelme save a json file for each image, the json file contains Label of each polygon annotated in the image, points which are multiple pixel value (x, y) of the whole polygon, shape_type of the object, which is polygon for our use case, image path and the image data in base64 value, and last imageHeight and imageWidth. But at the time of training the model accepts only one json file for test and training separately. That json file contains the key value pair of images which contains a further key value pair of height, width, id of the image and file name. Then the categories or we can say labels with their id, the labelling starts from 1 in mask RCNN because 0 is reserved for background. Further we have annotations for each image, which contains image id, bounding boxes and segmentation. All of these information was used to train the model efficiently.

c)Inferencing: Now after training the model, we tested it on our test dataset.

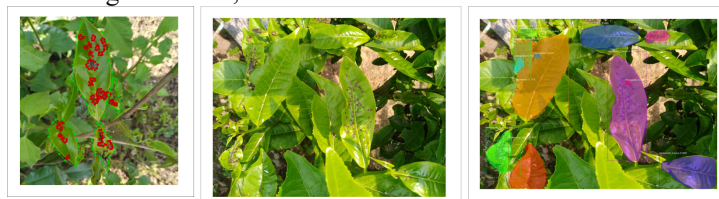

**Figure 16: labelling the image, Original Image and Predicted Image**

d)Evaluation: Further if we analyse both classes separately we found that how segmentation is done for each classes in the given image

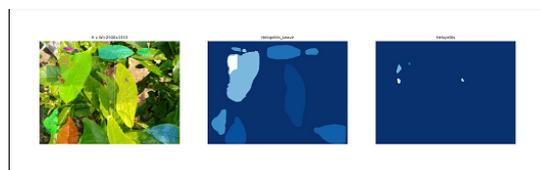

**Figure 17: Segmentation of tea leaf and disease**

Leave: From the above evaluation we can say that the model is performing well in case of detecting the diseased leave and also in segmenting them properly in case of overlapping cases On further analysing the we also found that the model is still having some difficulty in masking out the proper shape of the leaf. Which will affect the final results of calculating the whole leaf area.

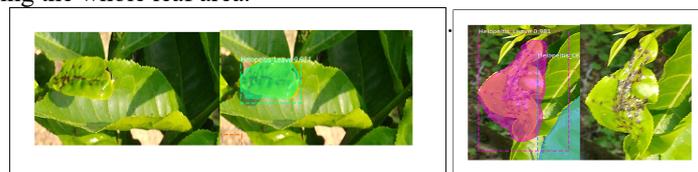

**Figure 18: Detection of disease leaf and Wrong detection of leaf shape**

So to improve these results we need more samples of images which consists of these kinds of complex shapes and also overlay samples to improve the model efficiency in detecting these complex cases.

Diseases: The model we trained now is not able to detect that much disease inside the images due to less data, but with more data we can make more improvements to the detection of disease in the leaves. It shows us that the disease part needs more focus.

e) Find The diseases belong to a single leave (Damage Area Calculation): Now to find the damage area of a leaf we need to first calculate the leaf part and then the damage part inside the leaf, and find the leaf damage percentage. The approach we used for now mainly focused on the ROI (region of interest) and Mask value of the model. The model on inference on an image returns 4 main values.

Mask: The binary value for each object in the image, which can be used for segmenting them.

ROI: The position of the image can be extracted using the ROI, it has four values [ y1, x1, y2, x2], we can use list slicing to extract the object from the whole image using these values, eg: image [ y1: y2, x1:x2].

Class Id: The class id for each object in the image.

Score: Classification score of the object

ROI here only gives 4 values, when we compared the values of Leaf ROI with disease ROI of the leaf we found that we can compare the values of both to find some pattern between them, but on further analysis we found using the four values won't solve the problem. So we calculated the coordinates of the diseases and leaf using the ROI values.

ROI [1032, 2139, 1962, 2550] => [y1, x1, y2, x2]

Coordinates (2139, 1032), (2550, 1032), (2550, 1962), (2139, 1962)

Coordinates (x1, y1), (x2, y1), (x2, y2), (x1, y2)

Now after calculating all the coordinates of the leaf and the diseases, we compared each coordinate x and y axis with disease x and y axis.

leave_x1 < disease_x1 and leave_y1 < disease_y1
leave_x2 > disease_x2 and leave_y1 < disease_y1
leave_x2 > disease_x2 and leave_y2 < disease_y2
leave_x1 < disease_x1 and leave_y2 > disease_y2

If all these conditions are met, we can say that the disease belongs to this leaf only. With the help of this we are able to find each individual disease and leaf which belongs to each other in the whole image

f) Calculate the total percentage of disease in the leave (Damage Area Calculation): To calculate the damage area inside the leaf we used the approach of counting the pixels of the image. For this approach first we used the Mask value returned by the model. The Mask value is in the shape of [height, width, total objects found], For each object found in the image we get a mask value for it, which tells us about the polygon shape of the object. The value of the Mask value is in true and False value. True for the pixels where the object is found. We first converted these mask values into int type 0 and 1 and replaced the value of 1 with 255 to plot a black and white image of the leaf and diseases. To get the image for the disease part we used the above approach first to find the disease of a leaf and then added all the mask numpy arrays of the disease which are inside the leaf, which return a single numpy array, which can be plotted. Now to get the area of disease and leaf, we calculated the white pixels in the image only. And then we just divided the area of disease with the leaf (area_disease / area_leaf) * 100

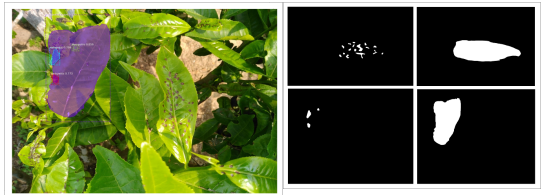

**Figure 19: Masking diseases under a single leaf and Visualizing disease spots and leaves**

## 4. CONCLUSION

As we know India is the second largest producer and exporter of tea in the world. And therefore it's one of the major crops which must be protected. Many machine learning and deep learning approaches are being made to detect disease in other plants. So we thought of an approach to protect our own major crop Tea. So, in this project we have evaluated 2 object detection models for the tea leaves disease detection part. The Mean Average Precision (mAP) of both the models trained for 10000 steps show that Faster R-CNN has 25% mAP and the SSD has 20% only. Since SSD mainly works for speed detection tasks by sacrificing accuracy, Faster R-CNN focuses on accuracy rather than speed. SSD can also provide good accuracy but since our task is related to detecting small disease parts in leaves therefore in some cases SSD fails to detect the disease parts more precisely. For the disease area calculation part, we have used MASK RCNN for segmentation tasks of diseases and diseased leaves. And then we calculated the disease area with the help of ROI and calculated all the coordinates of leaves and the disease in an image. Further we can improve the models with more disease data in tea leaves with different climate conditions and also other parts of the tea plant which can be attacked by diseases. The major challenges were to detect diseases in a large field, which can be done using drones and a high pixel camera which can enlarge the images of leaves to detect diseases.

## 5. ACKNOWLEDGMENT

The success and final outcome of this project required guidance and assistance from many people and we are extremely privileged to get this all along the completion of our project. We respect and thank our project coordinator Dr. Sajal Saha and Dr. Purnendu Bikash Acharjee and giving us all support and guidance. We are thankful to The Assam Agriculture University Dr. Popy Bora(Scientist) and Dr. B.C Nath(Assistant professor) to help us identify the disease. We are  thankful to all the teammates who work hard to accomplished the task.